\documentclass[11pt]{article}

\usepackage[]{ACL2023}

\usepackage{times}
\usepackage{latexsym}
\usepackage{utfsym}
\usepackage{pifont}
\usepackage[T1]{fontenc}
\usepackage[utf8]{inputenc}

\usepackage{microtype}
\usepackage{inconsolata}
\usepackage{times}
\usepackage{latexsym}
\usepackage{graphicx}
\usepackage{amsmath}
\usepackage{amsthm}
\usepackage{booktabs}
\usepackage{algorithm}
\usepackage{algorithmic}
\usepackage{subfigure}
\usepackage{multirow}
\usepackage{color}
\usepackage{bm}
\usepackage{xcolor}
\usepackage{amssymb}

\definecolor{G}{RGB}{45, 179, 0}
\definecolor{R}{RGB}{255, 102, 0}
\usepackage{graphicx}

\title{Divide, Conquer, and Combine: Mixture of Semantic-Independent \\Experts for Zero-Shot Dialogue State Tracking}

\author{Qingyue Wang\textsuperscript{$\spadesuit$$\clubsuit$}, Liang Ding\textsuperscript{$\diamondsuit$}, Yanan Cao\textsuperscript{$\spadesuit$\thanks{\ \ Yanan Cao is the corresponding author.}}, Yibing Zhan\textsuperscript{$\diamondsuit$}, Zheng Lin\textsuperscript{$\spadesuit$}, \\\textbf{Shi Wang\textsuperscript{$\heartsuit$}}, \textbf{Dacheng Tao\textsuperscript{$\nabla$}\and Li Guo\textsuperscript{$\spadesuit$}}\\
$\spadesuit$ Institute of Information Engineering, Chinese Academy of Sciences, Beijing, China \\
$\clubsuit$ School of Cyber Security, University of Chinese Academy of Sciences,
Beijing, China \\
$\heartsuit$ Institute of Computing Technology, Chinese Academy of Sciences, Beijing, China\\
$\diamondsuit$ JD Explore Academy, JD.com Inc, China  \ $\nabla$The University of Sydney, Australia\\
\texttt{
\{wangqingyue,caoyanan,linzheng,guoli\}@iie.ac.cn, wangshi@ict.ac.cn}\\
\texttt{
\{liangding.liam,zhanybjy,dacheng.tao\}@gmail.com
}}

\begin{document}
\maketitle
\begin{abstract}
Zero-shot transfer learning for Dialogue State Tracking (DST) helps to handle a variety of task-oriented dialogue domains without the cost of collecting in-domain data. Existing works mainly study common data- or model-level augmentation methods to enhance the generalization but fail to effectively decouple the semantics of samples, limiting the zero-shot performance of DST.
In this paper, we present a simple and effective ``divide, conquer and combine'' solution, which explicitly disentangles the semantics of seen data, and leverages the performance and robustness with the mixture-of-experts mechanism. Specifically, we divide the seen data into semantically independent subsets and train corresponding experts, the newly unseen samples are mapped and inferred with mixture-of-experts with our designed ensemble inference.
Extensive experiments on MultiWOZ2.1 upon the T5-Adapter show our schema significantly and consistently improves the zero-shot performance, achieving the SOTA on settings without external knowledge, with only 10M trainable parameters\footnote{Code is freely available at: \url{https://github.com/qingyue2014/MoE4DST.git}}.
\end{abstract}

\section{Introduction}
Dialogue state tracking (DST) plays an important role in many task-oriented dialogue systems \citep{Young2013POMDPBasedSS}.
The goal of this task is to understand users' needs and goals by exacting dialogue states at each turn,
which are typically in the form of a list of slot-value pairs~\citep{Wu2019TransferableMS}. Accurate DST performance can help downstream applications such as dialogue management.

However, collecting and annotating the dialogue state is notoriously hard and expensive \citep{Budzianowski2018MultiWOZA}.
This problem becomes pressing from single-domain to multi-domain scenarios.
To train a multi-domain DST model, dialogue annotators need to indicate all slot-value pairs for each domain and turn.
Therefore, tracking unseen slots in a new domain without any labels, i.e. zero-short prediction, is becoming an urgent demand for real-world deployments.

To make the DST module more practical, e.g. robust to unseen domains, various methods have been developed to improve the zero-shot capacity from the data-level or model-level. 
The first is to synthesize new dialogue samples or introduce other large labeled datasets (e.g QA datasets) to overcome the data scarcity issue~\cite{Campagna2020ZeroShotTL,li2021coco,Shin2022DialogueSA}.
\begin{figure}[t]
  \centering
  \includegraphics[scale=0.8]{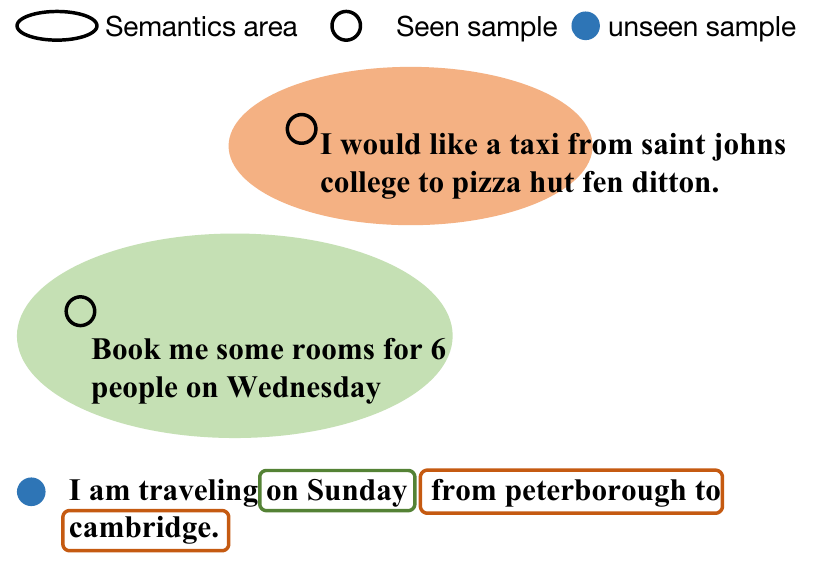}
  \caption{An illustration of the semantics areas of seen data and perform inference on the newly unseen sample. For each sample, We omit previous turns and only show the current utterance from the user. }
  \label{fig:motivation}
\end{figure}
The second line of work is to develop the advanced model/ framework to improve the scalability of DST, such as span-based approach, copy-augmented decoder, or pre-trained language model~\cite{chao2019bert,Wu2019TransferableMS,Wang2022SlotDM,zhong2023can}.
While empirically successful, we argue that the above data- or model-level augmentation methods have not explored the essence of zero-shot generalization, due to the lack of semantical disengagement ability to map the unseen sample to the seen data manifold~\cite{lazaridou2015hubness,li2017zero}.

To intuitively explain how the semantic areas of seen samples help in inferring the new unseen sample, we give an example in Figure~\ref{fig:motivation}. For an unseen sample from train domain, the \textit{\textcolor{G}{booking rooms}} area can help predict unseen slot ``train-day'', and the \textit{\textcolor{R}{booking a taxi}} area also help predict slot ``train-departure'' and ``train-destination''. As seen, a new unseen sample may be hard to directly infer due to the compositional complexity but can be easy to handle if mapped to related semantic-independent areas. But the representation-level disentanglement is challenging and unstable, especially for situations that require accurate semantic dividing.

In response, we provide a simple yet effective ``divide, conquer and combine'' solution to navigate the unseen sample to correspondingly accurate semantic experts. The philosophy is to explicitly divide the seen data into different semantic areas and train corresponding experts, and such data-level disentanglement provides flexibility to map the unseen sample to different semantic experts. The final output from the mixture-of-experts is expected to improve the zero-shot performance.
In practice, we design a three-step framework, where stages 1\&2 are for training and stage 3 is for inference: \ding{182}dividing: encode and cluster the semantics of seen data into subsets, \ding{183}conquering: train expert for each subset with dialogue state labels, and \ding{184}combining: mine the relationship between newly unseen sample and seen semantics, and perform ensemble inference with weighted experts.

Experimentally, we implement our framework upon T5-Adapter and demonstrate the effectiveness and universality of our proposed schema. 
Specifically, we achieve averaging 5\%$\sim$10\% improvement on the MultiWOZ benchmark with negligible training and deployment costs, achieving state-of-the-art zero-shot performance under settings without external information. Comprehensive analyses are reported to provide some insights to better understand our method.

\begin{figure*}[htbp]
  \centering
  \includegraphics[scale=0.46]{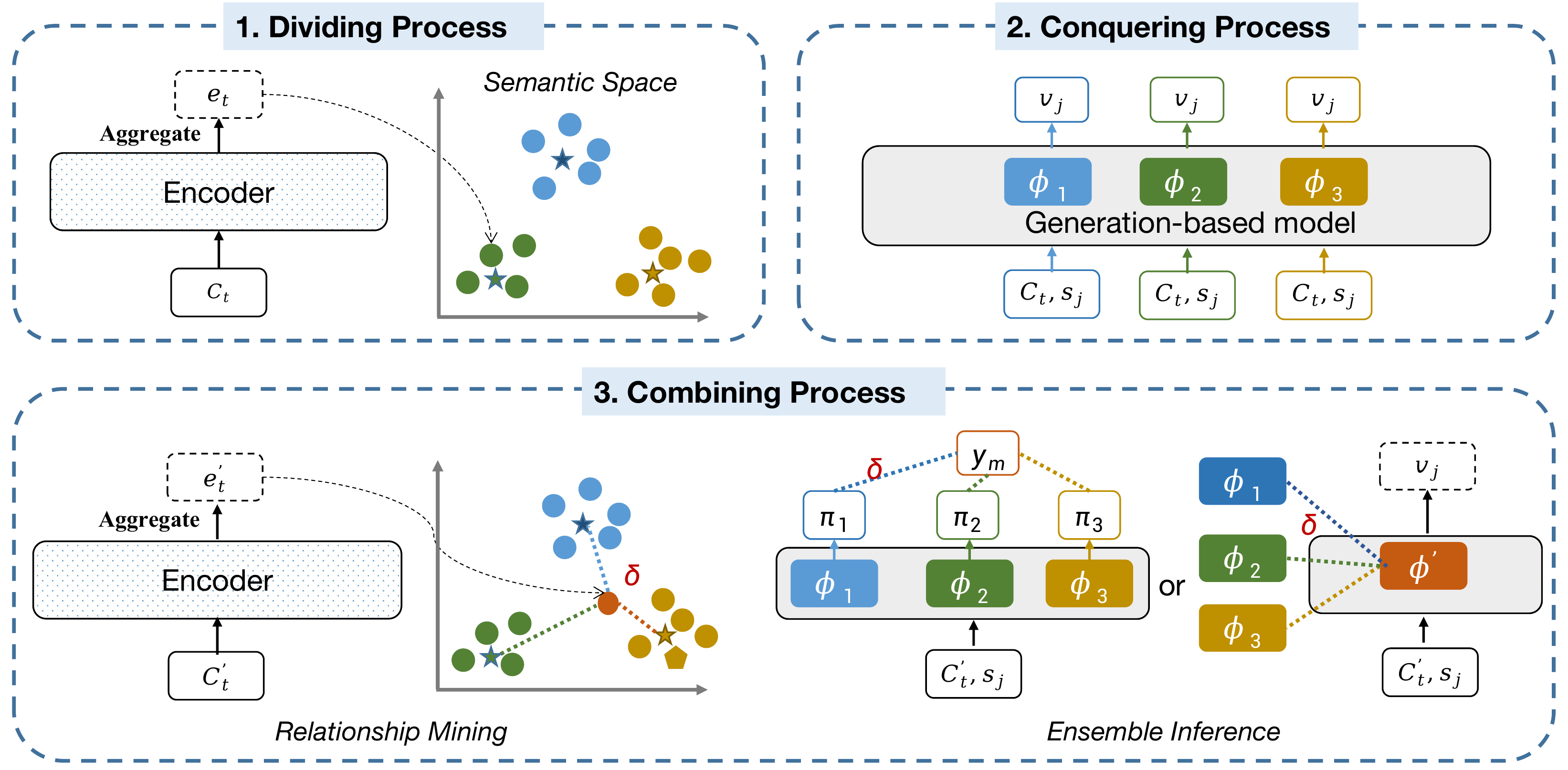}
  \caption{Illustration of our proposed schema (best viewed in color). }
  \label{fig:overview}
  \end{figure*}
\section{Related Work}
Dialogue State Tracking (DST) has been of broad interest to the dialogue research community. Existing DST models require plenty of state labels \citep{Henderson2014WordBasedDS, Zhong2018GlobalLocallySE,wu2020slotrefine}, which is hard to get in real scenarios.
Various studies on DST with zero-shot learning have been conducted to tackle unseen slots \citep{Yang2022PromptLF, Wang2022SlotDM} from the data or model perspective. Firstly, data augmentation is widely used to improve the effectiveness of the existing DST models. \citet{Campagna2020ZeroShotTL} synthesizes dialogues for a new domain using domain templates derived from observing a small dataset and the ontology of the domain. Other studies utilize diverse labeled datasets from other tasks, such as dialogue summarization task \citep{Shin2022DialogueSA} or generative question answering task \citep{Lin2021ZeroShotDS}, also called zero-shot cross-task transfer. In this paper, we focus on zero-shot cross-domain DST, where the model is first trained on several domains and transferred into unknown domains.

Many works focus on developing the advantage model or framework to enhance the robustness of DST~\citep{Wu2019TransferableMS,Kumar2020MADSTMB,wu2021bridging}. 
\citet{chao2019bert} adopts the Bert to produce context representations of dialogue context and applies span prediction modules to predict the slot value as a text span. \citet{Wu2019TransferableMS} encodes the whole dialogue context and decodes the value for every slot using a copy-augmented decoder.
Recently, many pre-trained language models, such as GPT \citep{radford2018improving} and T5 \citep{Raffel2019ExploringTL}, demonstrate impressive zero-shot learning ability and attract many researchers. \citet{Friedman2021SingledatasetEF} proposes to model multi-dataset question answering with a collection of single-dataset experts -- dataset-specific adapter modules \citep{Houlsby2019ParameterEfficientTL}. In DST,~\citet{Lin2021LeveragingSD} first leverages the slot description as a prompt and generates the slot value for zero-shot cross-domain settings. \citet{Wang2022SlotDM} models three types of slot dependency based on prompt learning and further improves the zero-shot performance. But these approaches mainly benefit from the similarity across slots and language knowledge inside pretrained models, ignoring the different semantics areas of seen data and failing to the effective inference on unseen domains.

\section{Background}
\textbf{Notation.}
We define $\{(A_1, U_1),.., (A_T, U_T)\}$ as a set of utterances from two speakers,
where $A$ and $U$ represent the system response and user utterance, respectively. 
At turn $t$, we denote the dialogue context as $C_t=\{(A_{1},U_{1}),\dots,(A_{t}, U_{t})\}$, which includes $t$ turns from system and user.
The task of DST is to predict the dialogue state $B_t$ given dialogue context $C_t$.
The dialogue state, $B_t$, is represented as slot-value pairs, denoted as $B_t=\{(s_1,v_1),\dots, (s_J, v_J)\}$ 
where $s_j$ and $v_j$ denote the $j$-th slot name and value at turn $t$. 
$J$ is the total number of slots in all domains.

\noindent \textbf{Generation-based DST.}
Unifying the dialogue states tracking as generation task shows promising performance, where it follows an auto-regressive fashion \citep{Lin2021LeveragingSD,Lee2021DialogueST}.
For each turn, a pre-trained language model (e.g T5) takes the dialogue context $C_t$ and the slot name $s_j$ as input and decodes the corresponding slot value $v_j$.
The objective $\mathcal{L}$ is to minimize the negative log-likelihood loss on all slots:
\begin{equation}
  \mathcal{L} = -\sum_{j=1}^JlogP(v_j|C_t, s_j)
\end{equation}
\section{Methodology}
\paragraph{Overviews}
Figure~\ref{fig:overview} illustrates the overview of our method following three steps. In the \ding{182}dividing process, a context encoder $f$ encodes seen dialogue contexts into representations to construct semantic space $\mathcal{E}$. These samples are then divided into several sub-sets by clustering. After that, We train semantic-independent DST experts using labeled states of sub-sets, also called the \ding{183}conquering process. During \ding{184}combining, we first estimate the relationships $\delta$ between seen data and unseen sample $C'_t$, and perform the weighted mixture-of-experts inference conditioned on $\delta$ for the unseen sample.

\subsection{Dividing Process}
The goal of data division is to obtain (ideally) semantic-independent areas for seen data. 
Previous works have shown that semantic disenchanted representation effectively improves the zero-shot generalization in the CV~\cite{Chen_2021_ICCV,ye2021disentangling} and NLP fields~\cite{shawetal2021compositional, furrer2020compositional}, but it's under-explored in dialogue, and also, we argue that data-level explicit dividing is simple and more interpretable than that of implicit representation-level dividing.

For the dialogue context, the division should consider multiple features, including domains, intentions of speakers even keywords of utterances, which is not feasible and costly in real scenarios.
We, instead, use the easy-to-use clustering algorithm, e.g. Kmeans~\cite{Hartigan1979AKC}, to achieve the sub-set dividing, where the pretrained contextual encoder~\cite{kenton2019bert,Raffel2019ExploringTL,vegav2,vegav1}, e.g. BERT and T5, is employed to accurately estimate the sample representation.

Specifically, given a dialogue context $C_t$, a context encoder $f$ is firstly applied to convert $C_t$ into the vector $e_t = {\rm Agg}[f(C_t)]$ in semantic space $\mathcal{E}$, where ${\rm Agg}$ is an aggregation operation (e.g. mean pooling). Afterward, we assign each context vector to one of the sub-sets by clustering algorithms:
\begin{equation}
  \mathcal{D}_k = {\rm clustering}(e_t), k\in\{1, ..., K\},
\end{equation}
where $\mathcal{D}_k$ represents the sample set of $k$-th sub-set and $K$ is the total number of sub-sets.

\subsection{Conquering Process}
In the conquering stage, sub-sets obtained in \ding{182}dividing process are used to train semantic-independent experts, respectively. In practice, we adopt a generation-based backbone model to model the DST task, and
the DST expert is trained with the samples of $k$-th sub-set :
 \begin{equation}
  \mathcal{L} = -\frac{1}{N_k}\sum_{n=1}^{N_k}\sum_{j=1}^{J}logP(v_j|C_t, s_j;\phi_k),
\end{equation}
where $N_k$ is the number of samples in $D_k$ and $\phi_k$ represents the parameters of $k$-th adapter. 
To benefit from the knowledge inside pre-trained models and avoid over-fitting on a single sub-set, we adopt T5~\cite{Raffel2019ExploringTL} as the generation backbone and only tune the corresponding adapter~\cite{Houlsby2019ParameterEfficientTL} for each expert.

\subsection{Combining Process}
\paragraph{Relationship Mining} 
Given an unseen sample, we map its dialogue context $C'_t$ under space $\mathcal{E}$ to obtain the semantic vector $e'_t$ (i.e., $e'_t={\rm Agg}[f(C'_t)]$). 
Then, the relationship between semantic areas and the unseen sample is computed by:
\begin{equation}
\delta (C'_t, \mu_k) = \frac{\exp(d(e'_t,\mu_k)/\tau)}{\sum_{k=1}^{K}\exp(d(e'_t, \mu_k)\tau)},
\label{eq:temp}
\end{equation}
where $d$ is a distance function and $\tau$ is a scalar temperature. $\mu_k$ is the prototype of a semantic area by averaging all vectors of samples in $\mathcal{D}_k$.
\paragraph{Ensemble Inference}
We consider two ensemble strategies that are widely used in AI challenges~\cite{ding2019usyd,ding2021usyd} to realize the relation-based mixture-of-experts inference, also denoted as ensemble inference: \textit{parameters-level} and \textit{token-level}. (1) Parameter-level ensemble initializes a new adapter $\phi'$ using the weighted sum parameters of trained-well adapters $\{\phi_k\}_{k=1}^{K}$: 
\begin{equation}
    \phi' = \sum_{k=1}^{K}\delta(C'_t, \mu_k)\phi_k
\end{equation}
And then, the model returns the prediction with the maximum probability under $P(v_j|C'_t,s_j; \phi')$.
(2) Token-level ensemble combines the prediction of trained-well experts to generate one sequence step by step. Formally, we generates the $m$-th target token $y_m$ of value $v_j$ with a weighted sum prediction of adapters:
\begin{equation}
  \begin{split}
    \pi_k &= log P(w|y_{(<m)},C'_t, s_j; \phi_k),\\
    y_{m} & = \mathop{\mathrm{argmax}}\limits_{w\in \mathcal{W}}\sum_{k=1}^K \delta(C'_t, \mu_k)\cdot \pi_k 
  \end{split}
  \end{equation}
  where $\pi_k$ is the predicted word distribution when using adapter $\phi_k$. Notably, parameter-level ensemble inference, requiring deploying only a new single adapter, enjoys extremely low deployment costs, while token-level one owns the better model capacity and is expected to perform better.
  
\section{Experiments}
\paragraph{Dataset}
We evaluate our method on widely-used multi-domain datasets MultiWOZ \citep{Budzianowski2018MultiWOZA} and Schema-Guided Dataset  \citep{Rastogi2020TowardsSM}. The MultiWOZ dataset contains 10k+ dialogues across 7 domains.
Each dialogue consists of one or multiple domains. We follow the previous pre-processing and evaluation setup \citep{Lin2021LeveragingSD,Wang2022SlotDM}, where the restaurant, train, attraction, hotel, and taxi domains are used for zero-shot cross-domain experiments.
The Schema-Guided Dialogue (SGD) dataset consists of over 16k+ multi-domain dialogues and covers 16 domains. The test set contains unseen data to measure the performance in the zero-shot setting. Detailed data statistics are shown in Appendix~\ref{sec:appendix}.
% Appendix \ref{sec:appendix} gives the statistics of the MultiWOZ and SGD datasets.
\begin{table*}[htbp]
  \centering
 \resizebox{0.99\linewidth}{!}
 {
  \begin{tabular}{lllllllll}
    \toprule
    \multirow{2}{*}{\textbf{Model}} & \multirow{2}{*}{\textbf{\begin{tabular}[c]{@{}l@{}}\#Trainable \\ Parameters\end{tabular}}} & \multirow{2}{*}{\textbf{\begin{tabular}[c]{@{}l@{}}Pretrained-\\ model\end{tabular}}} & \multicolumn{6}{c}{\textbf{Joint Goal Accuracy}}                                                               \\
    \cmidrule(lr){4-9}
                                    &                                                                                             &                                                                                       & \textbf{Attraction} & \textbf{Hotel} & \textbf{Restaurant} & \textbf{Taxi} & \textbf{Train} & \textbf{Average} \\ \midrule
    TRADE \citep{Wu2019TransferableMS} & -                                                                                           & N                                                                                     & 19.87               & 13.70           & 11.52               & 60.58         & 22.37          & \underline{25.76}            \\
    MA-DST \citep{Kumar2020MADSTMB} & -                                                                                           & N                                                                                     & 22.46               & 16.28          & 13.56               & 59.27         & 22.76          & \underline{26.87}            \\
    SUMBT \citep{Lee2019SUMBTSM}  & 440M                                                                                        & Bert-base                                                                             & 22.60                & 19.80           & 16.50                & 59.50         & 22.50           & \underline{28.18}            \\
    T5DST \citep{Lin2021LeveragingSD} & 60M                                                                                         & T5-small                                                                              & 33.09               & 21.21          & 21.65               & 64.62         & 35.42          & \underline{35.20}             \\
    T5DST $^\dag$\citep{Lin2021LeveragingSD} & 220M                                                                                         & T5-base                                                                              & 35.51               & 22.48          & 25.04               &  65.93         & 37.82          & \underline{37.36}             \\
    SlotDM-DST \citep{Wang2022SlotDM} & 60M                                                                                       & T5-small                                                                          & 33.92 & 19.18                     &20.75                   & 66.25 & 36.96    & \underline{35.55} \\
    SlotDM-DST \citep{Wang2022SlotDM} & 220M                                                                                       & T5-base&37.83 &26.50 &27.05 &\bf 69.23 & 40.27 & \underline{40.18} \\
    TransferQA \citep{Lin2021ZeroShotDS} & 770M                                                                                       & T5-large                                                                          & 31.25 &22.72                       &26.28                   & 61.87 & 36.72    & \underline{35.77} \\
    \midrule  
    \multirow{2}{*}{T5-Adapter$^\dag$}     & 0.8M                                                                                        & T5-small                                                                              & 33.85               & 18.22          & 19.62               & 64.93         & 32.25          & \underline{33.77}                 \\
                                    & 3.6M                                                                                        & T5-base                                                                               & 39.98               & 23.28          & 28.58                & 65.03         & 36.98          & \underline{38.77}                 \\ \midrule\midrule
    Ours (Param-level)              & \multirow{2}{*}{0.8M$\times$$K$}                                                                   & \multirow{2}{*}{T5-small}                                                             & 34.63               & 24.22          & 22.07               & 65.41         & 33.88          & \underline{36.02}            \\
    Ours (Token-level)              &                                                                                             &                                                                                       & 35.82               & 24.78          & 22.86               & 65.87         & 40.27          & \underline{\textbf{37.92}}                 \\
    \midrule
    Ours (Param-level)              & \multirow{2}{*}{3.6M$\times$$K$}                                                                   & \multirow{2}{*}{T5-base}                                                              & 41.28               & 26.15          & 31.05               & 66.64         & 38.72          & \underline{40.76}                  \\
    Ours (Token-level)              &                                                                                             &                                                                                       & \textbf{41.35}               & \textbf{27.72}          & \textbf{33.76}               & 66.90         & \textbf{43.81}          & \underline{\textbf{42.71}}                 \\ \bottomrule   
  \end{tabular}}
  \caption{Zero-shot results on MultiWOZ 2.1 dataset. All numbers are reported in joint goal accuracy (\%) and the best results among each setting are bolded.
  $K$ 
  is a hyper-parameter and refers to the number of sub-sets. Expect for $^\dag$, all results of baselines come from the original papers.}
  \label{tab:zero}
  \end{table*}
\paragraph{Evaluation Metrics}
We follow~\citet{Lin2021LeveragingSD} to use slot accuracy (SA) and joint goal accuracy (JGA) as evaluation metrics. SA is calculated as the ratio of individual slot in which its value is correctly predicted, and JGA measures the percentage of correct in all dialogue turns, where a turn is considered as correct if and only if all the slot values are correctly predicted. In zero-shot DST \citep{Wu2019TransferableMS, Lin2021LeveragingSD}, the model obtains all training data from the training dialogues except for an unseen domain, which is used to evaluate.

\paragraph{Comparison Baselines}
We evaluate our model against existing zero-shot DST baselines. \textbf{TRADE} \citep{Wu2019TransferableMS} utilizes a copy mechanism to track slot values for unseen domains.
\textbf{MA-DST} \citep{Kumar2020MADSTMB} designs multiple layers of cross-attention to capture relationships at different levels of dialogue granularity.
\textbf{SUMBT} \citep{Lee2019SUMBTSM} proposes a non-parametric method to score each candidate slot-value pair in a pre-defined ontology.
\textbf{TransferQA} \citep{Lin2021ZeroShotDS} is a cross-task zero-shot DST method where the model is pre-trained on QA datasets and then applied to unseen domains.
\textbf{T5DST} \citep{Lin2021LeveragingSD} explores the slot description as a prompt to generate slot values.
\textbf{SlotDM-DST} \citep{Wang2022SlotDM} models three types of slot dependency, i.e., slot-slot, slot-value, and slot-context, to improve zero-shot DST.
\textbf{SGD-baseline} utilizes schema descriptions to predict the dialogue state of unseen domains.
Moreover, we implement \textbf{T5-Adapter} that concatenates the dialogue context and slot name as inputs, following T5DST, as the fair baseline of our method.
Different from other baselines fine-tuning all parameters, T5-Adapter only tunes the parameters of the adapter during training. All baselines listed here do not consider any information from new domains. For a fair comparison, we don't include the in-context learning work on~\citet{Hu2022InContextLF} because they design specific prompts using the information from the unseen domain.

\paragraph{Implementation}
Our models are implemented in Pytorch \citep{Paszke2019PyTorchAI} using HuggingFace \citep{Wolf2019TransformersSN} and the adapter-transformers library \citep{Pfeiffer2020AdapterHubAF}. In division processing, we utilize T5-base \citep{Raffel2019ExploringTL} as the context encoder and apply mean pooling on the outputs of the encoder as the dialogue vectors.
We choose Kmeans \citep{Hartigan1979AKC} as the clustering algorithm and set the number of sub-sets as 3. 
In conquer processing, T5 is employed as the DST expert with the default adapter configuration from \citet{Houlsby2019ParameterEfficientTL}\footnote{Note that users could employ advanced Adapters or Prompts~\cite{he2022sparseadapter,zhong2022panda} to obtain better performance with fewer parameters, which will be explored in our future work.}, which adds approximately 0.8M parameters to the T5-small (60M) and 3.6M parameters to the T5-base (220M). We freeze the transformer parameters and use a learning rate of 1e-4 on adapter parameters for each expert. For all experiments, we train each independent expert for 10 epochs.  We use the AdamW optimizer \citep{Loshchilov2017DecoupledWD} and set the batch size to 16. In the combining process, the scale temperatures are set to 2 and 0.2 in the token- and parameter-level ensemble inference, respectively. 
For a fair comparison, we process and evaluate the MultiWOZ datasets following T5DST~\cite{Lin2021ZeroShotDS}. In the SGD dataset, we process the data following TransferQA~\cite{Lin2021LeveragingSD} and use the official evaluation script\footnote{\url{https://github.com/google-research/google-research/tree/master/schema_guided_dst}} to evaluate. 
% on which the model is evaluated.
  
\subsection{Main Results}
\label{subsec:mainresults}
\paragraph{Our method significantly improves zero-shot cross-domain performance.}
% \label{sec:zero-shot}
Table \ref{tab:zero} shows the zero-shot DST results on MultiWOZ 2.1 dataset.
Among these baselines, those methods using the T5 model have a much better performance than those without pre-trained models (e.g.TRADE), illustrating the strong transfer ability of pretrained models in zero-shot settings.
Interestingly, the T5-Adapter yields +1.41\% average over the fine-tuning on T5-base (T5DST), which has not been discussed in previous DST works, indicating that few trainable parameters are also effective in transfer learning.
Among all models, our method achieves state-of-the-art performance on average (42.71\%) with about 10M trainable parameters (when $K$=3). And there is a great improvement in the `train' domain.  The reason is that all slots in that domain are closely related to seen data, which easily benefits from the method we propose. Additionally, the token-level ensemble inference as expected obtains higher joint goal accuracy improvements than the parameter-level one across all domains. However, the token-level ensemble needs more computations during inference. Detailed analysis on ensemble inference is discussed in \S\ref{sec:ensemble}.

Table \ref{tab:sgd} shows the zero-shot performance on the SGD dataset. In the SGD dataset, there are four domains in the testing set but are not in the training set. So we train the proposed model using the whole training set and test on these four unseen domains for the zero-shot setting. Compared with the SGD baseline, the zero-shot performance of our model is consistently higher in four unseen domains. 
 \begin{table}[t]
 \centering
 % \scalebox{0.65}
 \resizebox{\linewidth}{!}{
 \begin{tabular}{lllll}
 \toprule
 \textbf{Domain} & \textbf{SGD-baseline} & \textbf{TransferQA} & \textbf{Seq2seq-DU} & \textbf{Ours} \\ \midrule
 Messaging & 10.2 & 13.3 & 4.9 & 28.7/22.1 \\
 Payment & 11.5 & 24.7 & 7.2 & 19.4/19.1\\
 Trains & 13.6 & 17.4 & 16.8 & 42.3/40.6 \\
 Alarm & 57.7 & 58.3 & 55.6 & 68.8/68.7 \\ \midrule
 Average &20.5 & 25.9 &20.3 & \textbf{39.8}/37.6\\ \bottomrule
 \end{tabular}}
 \caption{Zero-Shot results on SGD dataset. All results are reported in JGA (\%). Our results are listed under the token-level/parameter-level ensemble.} 
 \label{tab:sgd}
 \end{table}
 
\paragraph{Our method also effectively enhances the full-shot performance.}
The philosophy of our mixture of semantic-independent experts has the potential to improve the full-shot settings. To validate our hypothesis, we conduct full-shot experiments and list the results in 
Table \ref{tab:full}.
As shown, our approach still shows superiority against the strong T5-Adapter baseline and other existing works, demonstrating the universality of our method.
\begin{table}[t]
  \centering
  % \scalebox{0.75}
  \resizebox{\linewidth}{!}
  {
  \begin{tabular}{llll}
  \toprule
  \textbf{Model} & \begin{tabular}[c]{@{}l@{}}\textbf{\#Trainable}\\ \textbf{Parameter}\end{tabular} & \begin{tabular}[c]{@{}l@{}}\textbf{Pre-trained}\\ \textbf{Model}\end{tabular} & \textbf{JGA}  \\ \midrule
  TRADE &- & N & 45.60 \\
  STARC \citep{Gao2020FromMR} &440M& Bert-base & 49.48 \\
  SGD-baseline &440M& Bert-base &43.40 \\
  T5DST & 220M & T5-base & 53.15 \\
  T5-Adapter &3.6M & T5-base & 52.14 \\
  \midrule
  Ours (Param-level) &3.6M$\times$$K$ & T5-base & 52.54 \\
  Ours (Token-level) &3.6M$\times$$K$ & T5-base & 54.35 \\ \bottomrule
  \end{tabular}}
  \caption{Full data results on MultiWOZ 2.1 dataset. For a fair comparison, only those generative models with the ability of zero-shot inference are listed here.}
  \label{tab:full}
  \end{table}

\section{Discussion}
To better understand our proposed schema, we first present essential \textbf{\em ablation} studies in \S\ref{subsec:ablation}, and show in-depth analyses on \textbf{\em clustering} (\S\ref{sec:cluster}) and \textbf{\em ensemble inference} (\S\ref{sec:ensemble}), respectively. Additionally, we discuss the \textbf{\em complementarity} of our framework with others in \S\ref{subsec:complmentary}.
\subsection{Ablation Study}
\label{subsec:ablation}
\begin{figure}[t]
 \centering
 \vspace{-13pt}
 \begin{minipage}[c]{0.23\textwidth}
 \centering
 \includegraphics[scale=0.25]{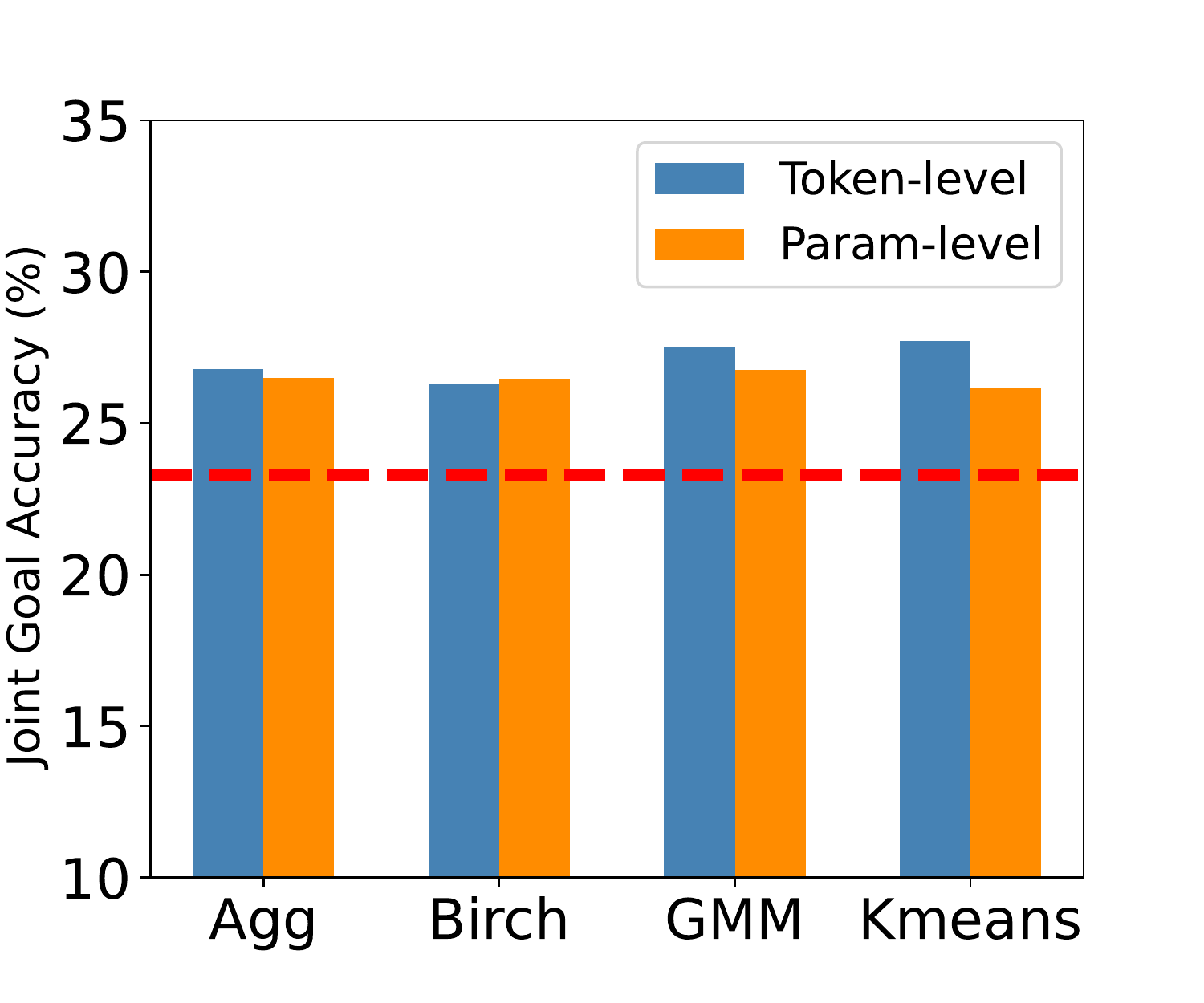}
 \caption{Impact of different clustering algorithms on hotel domain.}
 \label{fig:clu_algorithm}
 \end{minipage}
 \begin{minipage}[c]{0.23\textwidth}
 \centering
 \includegraphics[scale=0.25]{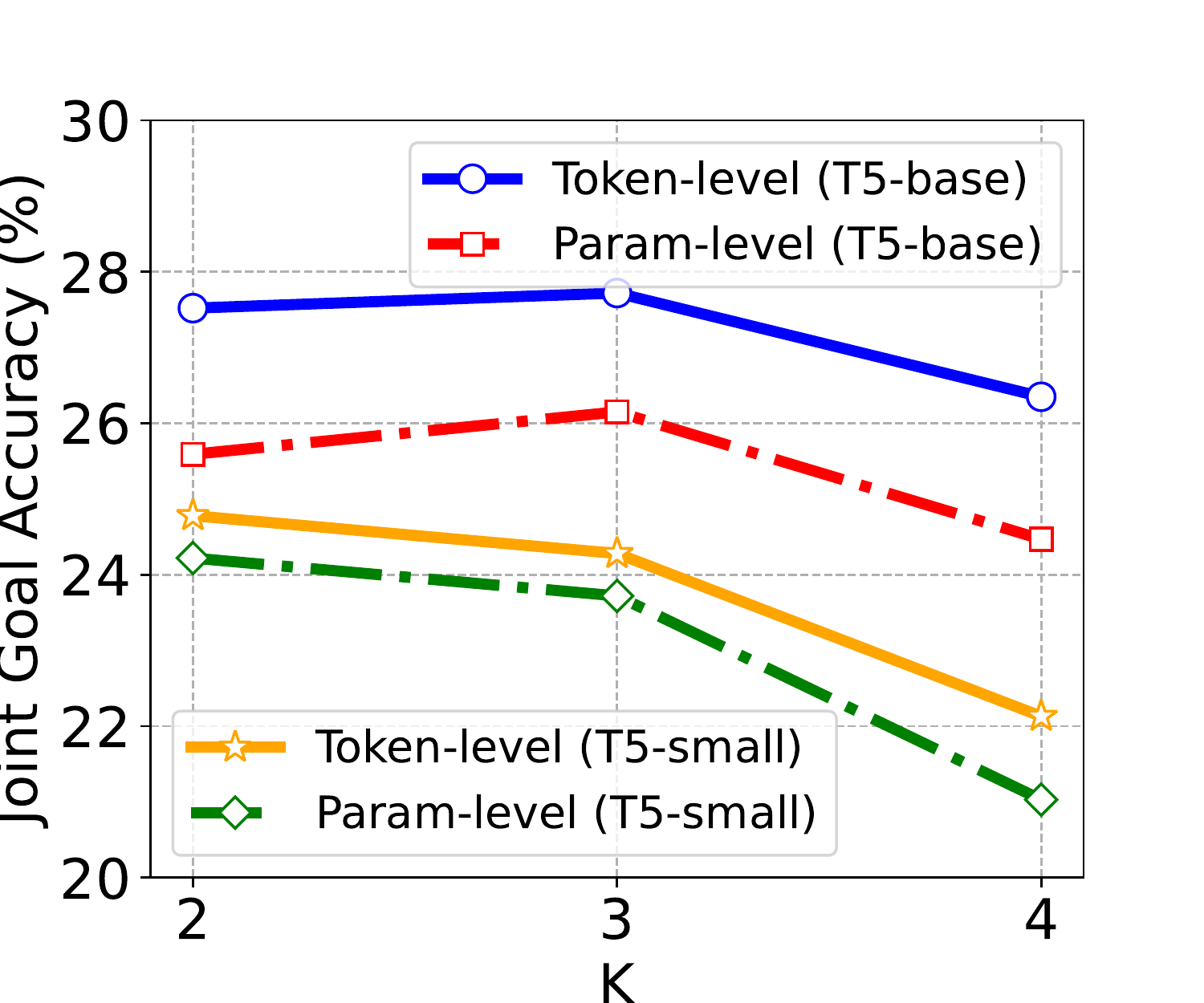}
 \caption{Impact of different numbers of sub-sets on hotel domain.}
 \label{fig:base_k}
 \end{minipage}
 \end{figure}

To understand the effects of major components, we conduct ablation studies on MultiWOZ 2.1 dataset.
\paragraph{Impact of Clustering Algorithms}
We study the effect of different clustering algorithms, including Kmeans~\cite{Hartigan1979AKC}, Birch~\cite{Zhang1996BIRCHAE}, Agglomerative~\cite{Gowda1978AgglomerativeCU}, and GMM~\cite{Yang2012ARE} on hotel domain in Figure~\ref{fig:clu_algorithm}. As shown, 1) all clustering algorithms perform better than the T5-Adapter (\textcolor{red}{Red} dotted line), showing the effectiveness and stability of our framework; and 2) GMM achieves the best performance on parameter-level ensemble inference while our chosen Kmeans wins on token-level ones. We believe advanced clustering may bring better division, thus achieving further improvement, which will be investigated in future work.
\paragraph{Impact of Number of Subsets}
We conduct experiments to observe the influence of the number of subsets during data division. Experiments on hotel domain with different $K$ values are in Figure \ref{fig:base_k}. We find that the joint goal accuracy performance increases with the value of $K$ first and then decreases on T5-base. The results show that the optimal number of sub-sets is 2 for T5-small and 3 for the T5-base model. Noted that our model strongly depends on the data distribution and data partition, which means that the zero-shot performance may not increase linearly as K increases. 

\begin{figure}[t]
  \centering
  \includegraphics[scale=0.25] {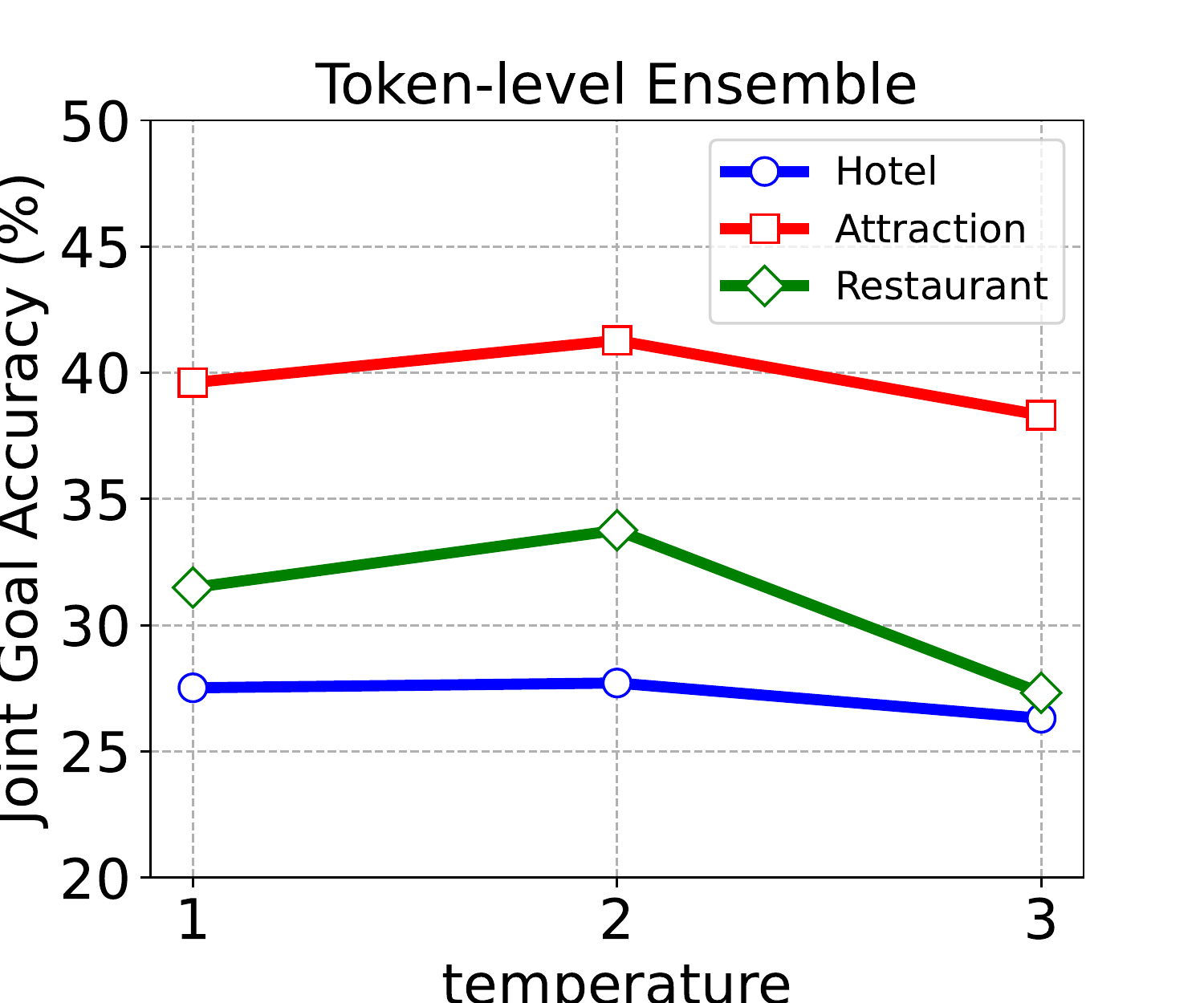}
  \hspace{-3mm}
  \includegraphics[scale=0.25] {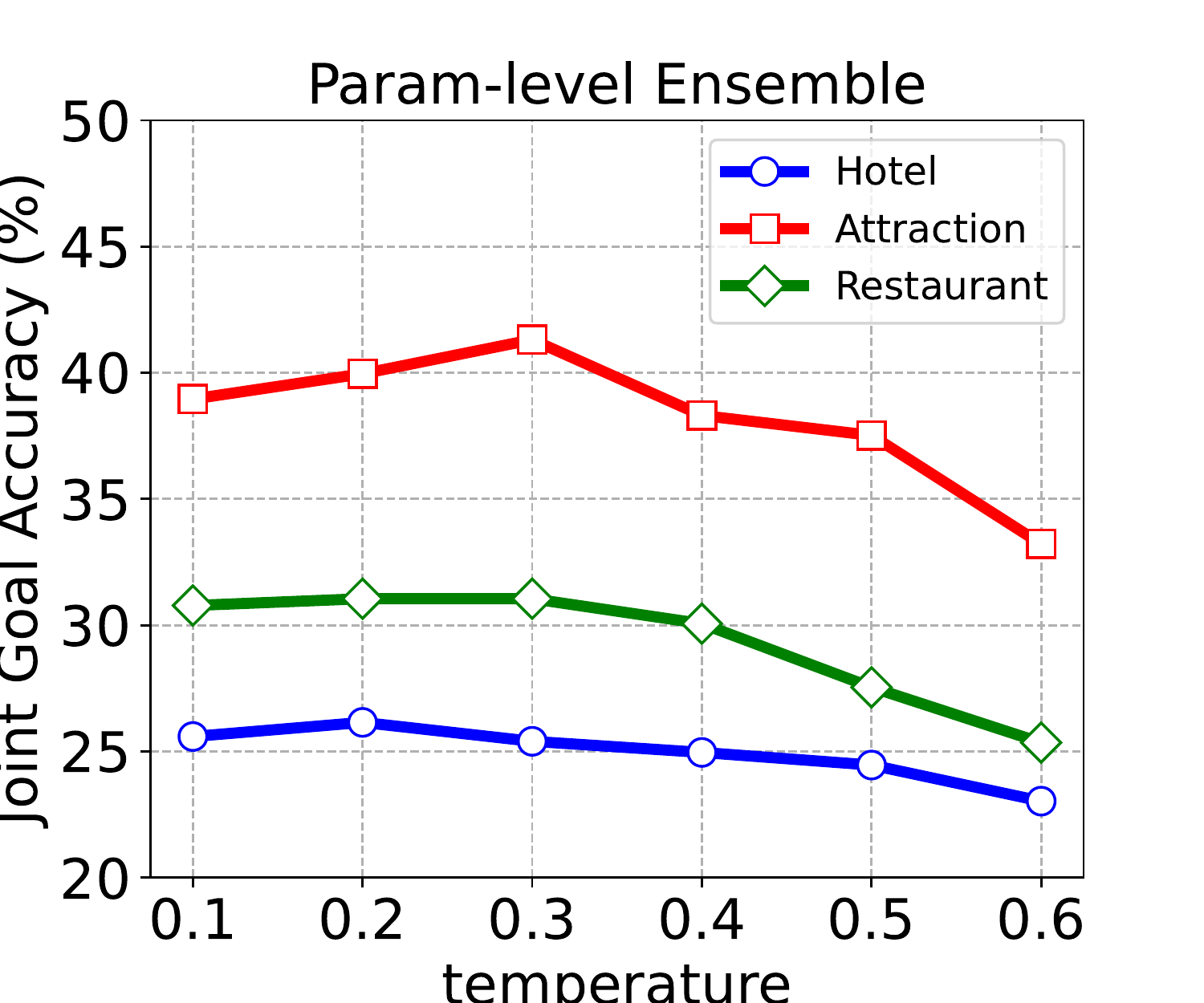}
  \caption{Impact of different temperatures $\tau$.}
  \label{fig:temp}
  \end{figure}
\paragraph{Impact of Temperature}
The scale of temperature in Equation~\ref{eq:temp} actually controls the smoothness of the weights and output distribution in the mixture of trained-well experts upon language models~\cite{peng2023towards}. As $\tau \to +\infty$, the weights become smoother. Contrarily, the distance collapses to a point mass when $\tau \to 0$. 
We study its influence on three domains in Figure \ref{fig:temp}. 
As shown, for token-level ensembles, larger temperature ($\ge1$) achieves better performance while smaller temperatures ($\leq0.4$) facilitate the parameter-level ensemble inference.
We suppose that the parameter space of semantic-independent experts is nearly orthogonal so that a smoother weight combination may hurt its performance. Differently, smoother weights are suitable for the token-level since the predictions from different experts are required to be easily merged. And the performances can be further improved by hyper-parameters searching.

\paragraph{Impact of Weight in Combining Process}
Mapping the unseen sample to existing subsets and obtaining the mapping weights are central in \ding{184}combing process. Besides adopting the weights by inference from the trained clustering model, we try other two weights: 1) \textit{argmax}: assigning 1 for the subset with max mapping probability and 0 for others, and 2) \textit{average}: assigning uniform probability for all subsets. As shown in Table~\ref{tab:ablation}, directly leveraging the inference weights shows the best performance for both parameter-level and token-level ensemble inference, showing the necessity of reusing the clustering model as the proxy for relationship mining.

\begin{table}[t]
  \centering
  % \scalebox{0.65}
  \resizebox{\linewidth}{!}
  {
    \begin{tabular}{lllll}
      \toprule
      \multirow{2}{*}{\textbf{Weights}} & \multicolumn{2}{c}{\textbf{Hotel}} & \multicolumn{2}{c}{\textbf{Taxi}} \\ \cline{2-5} 
                                      & Param-level            & Token-level            & Param-level           & Token-level           \\ \midrule
      Ours                            & 26.15            & 27.72            & 66.64           & 66.90\\
       Argmax                 & 24.47            & 24.85            & 65.09           & 66.38\\
      Average                     & 20.62            & 25.87            & 59.61           & 65.51 
                \\ \bottomrule
      \end{tabular}}
  \caption{The Impact of weight in combing process.}
  \label{tab:ablation}
  % \vspace{-10pt}
  \end{table}

\subsection{Analysis on Clustering}
\label{sec:cluster}
\begin{figure}[t]
  \centering
  \vspace{-18pt}
\includegraphics[scale=0.27] {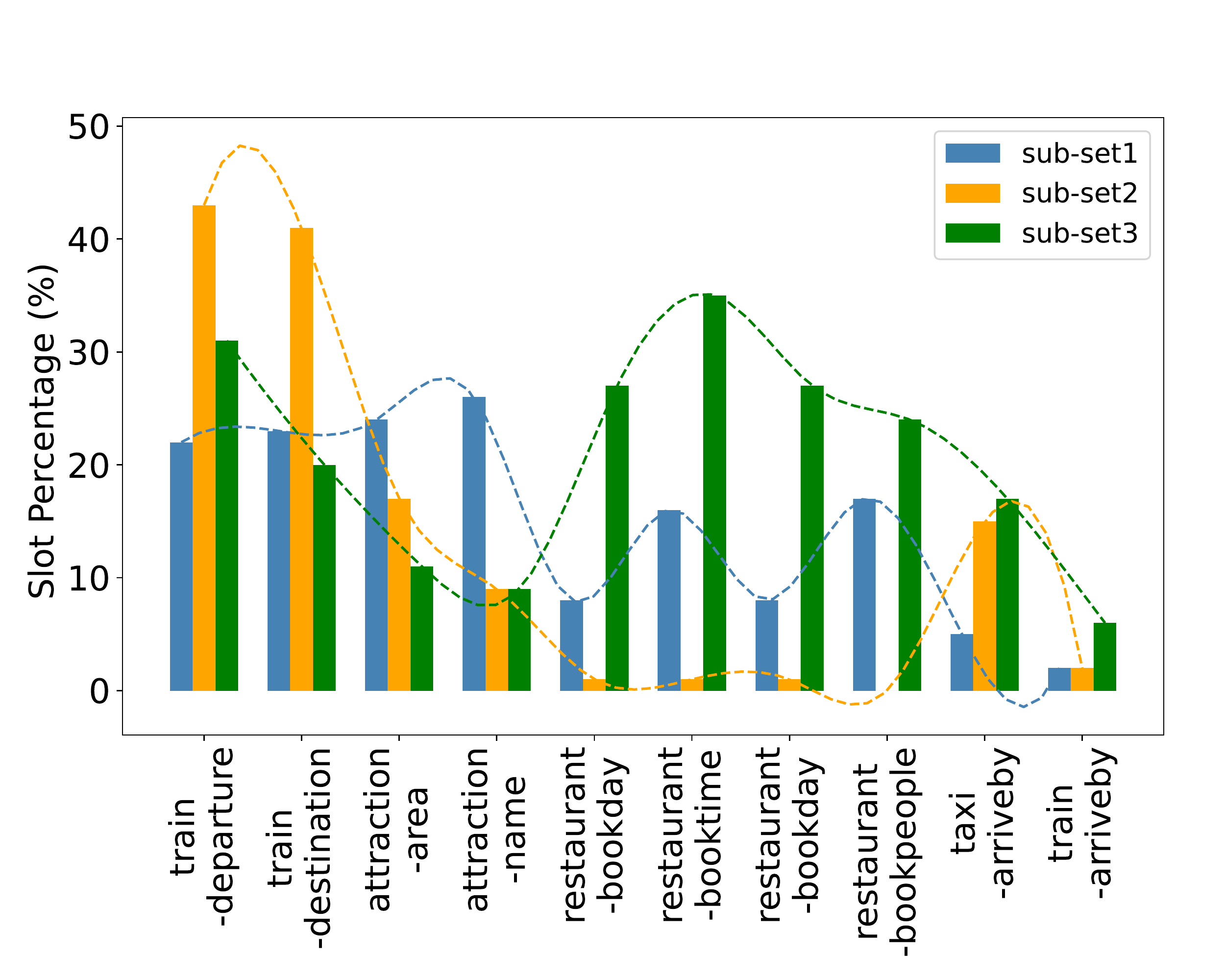}
\vspace{-10pt}
  \caption{Statistics of slot distribution across sub-sets.}
  \label{fig:cluster_dis}
  % \vspace{-10pt}
  \end{figure}

\begin{figure}[t]
  \centering
  \vspace{-10pt}
\includegraphics[scale=0.36
] {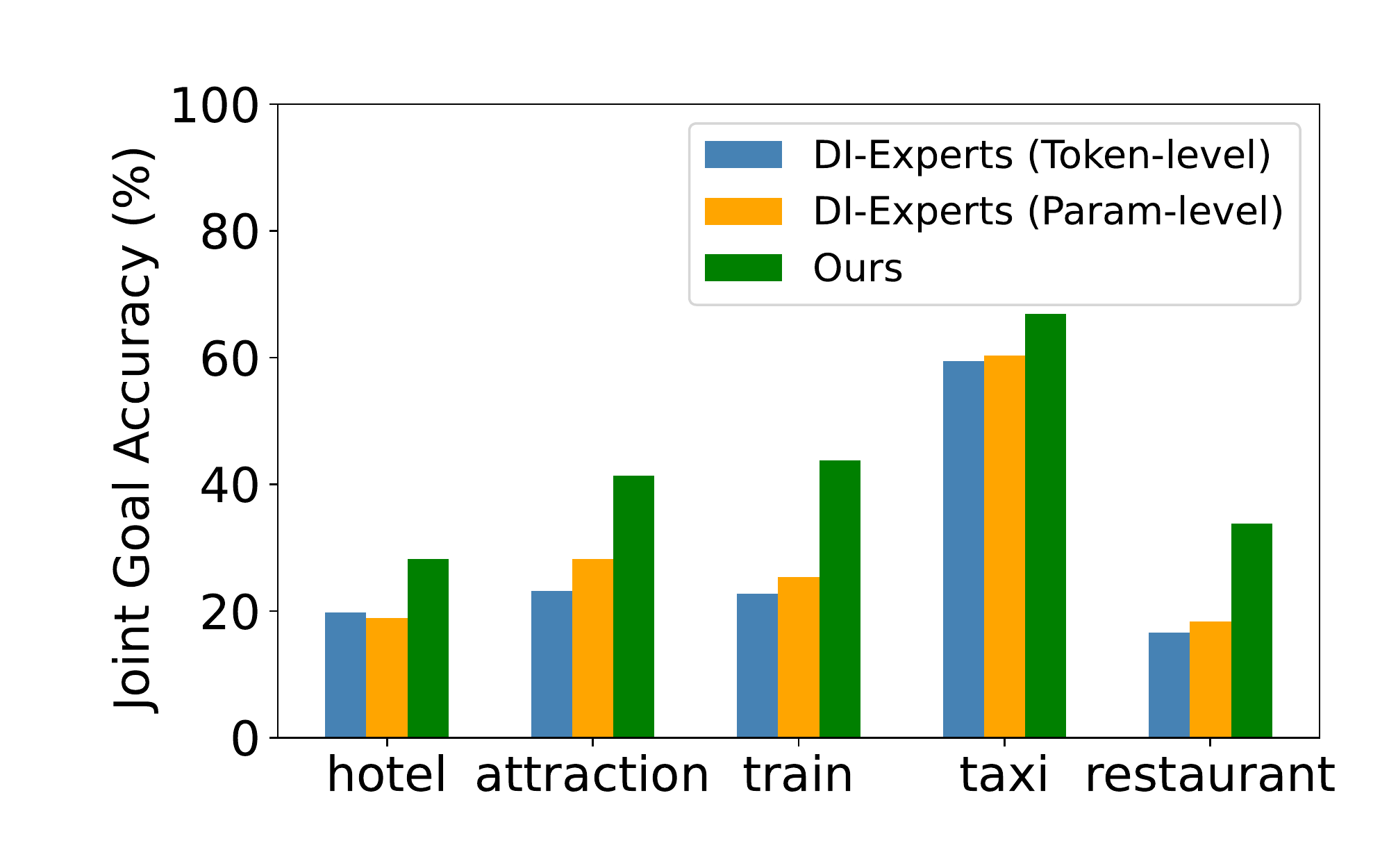}
\vspace{-10pt}
  \caption{The zero-shot performance of DI-Experts.}
  \label{fig:domain_per}
  \vspace{-10pt}
  \end{figure}

\paragraph{Robust to Different Context Encoders}
\begin{figure}[t]
  \centering
  \includegraphics[scale=0.027] {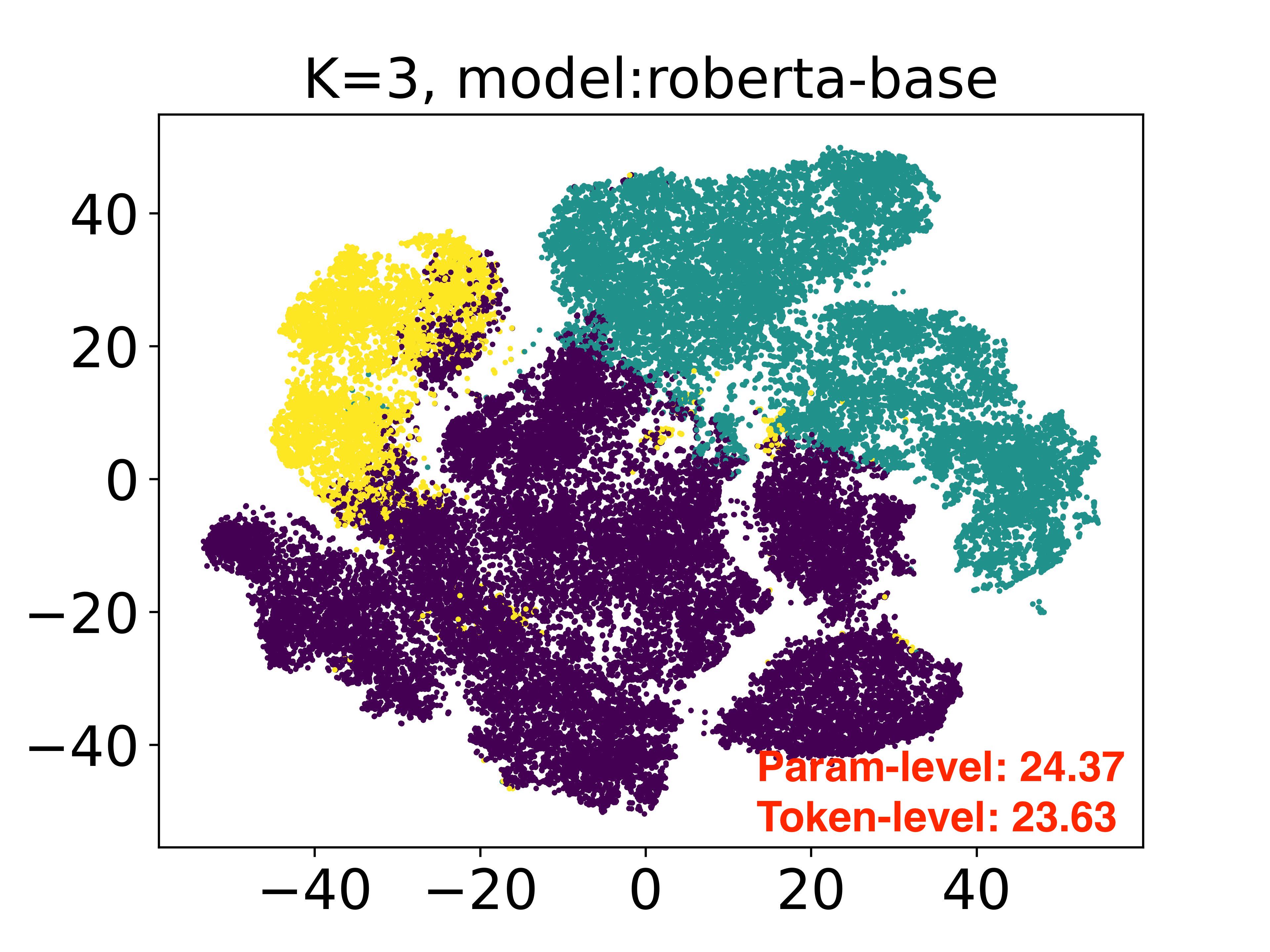}
  \hspace{-2mm}
  \includegraphics[scale=0.027] {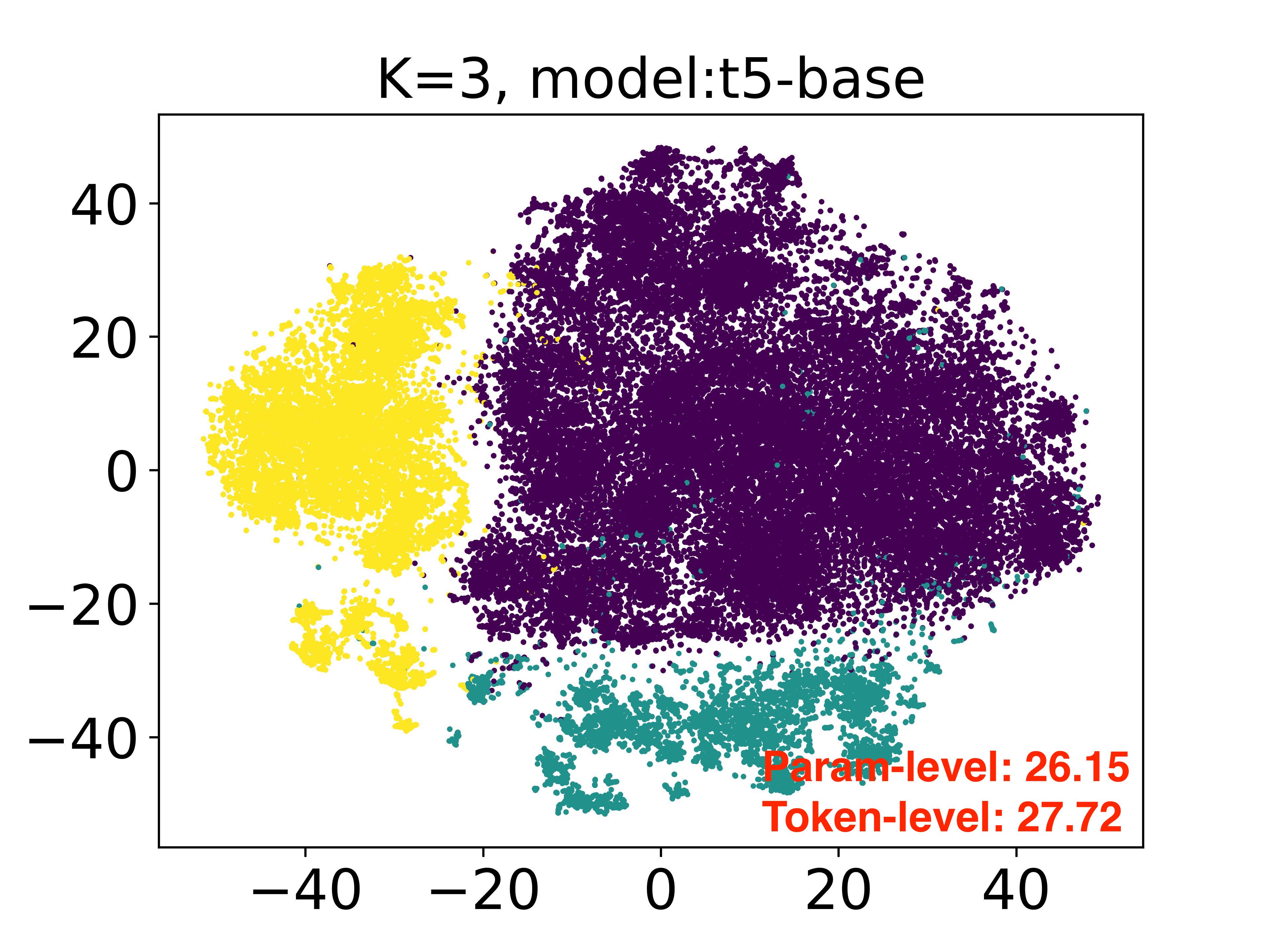}
  \caption{The t-SNE visualizations of clustered subsets represented with T5 and RoBERTa.
  % (except hotel domain)
  % when using different pre-trained models. 
  ``\textcolor{red}{Token-level}'' and ``\textcolor{red}{Param-level}'' show the zero-shot performance.}
  \label{fig:pretrain}
\end{figure}
To check whether the clustering method is robust to different context encoders, e.g. RoBERTa~\cite{Liu2019RoBERTaAR} and T5~\cite{Raffel2019ExploringTL}. 
We visualize their representation in Figure~\ref{fig:pretrain} with their corresponding zero-shot performance attached, and show that 1) both context encoders nicely represent the seen data and could map them to visually separated semantic areas, and 2) better context encoder, i.e. T5, indeed brings much clear semantic separate degree, thus leading to better zero-shot performance, i.e. T5$>$RoBERTa.
These findings confirm that clustering is simple, reasonable, and robust to different content encoders to obtain separate semantic areas.
\paragraph{Brings Explicit Semantic Division in Data}
To explicitly analyze the semantics division of clustered subsets, we randomly sample four hundred for each sub-set and compute the slot distribution in Figure~\ref{fig:cluster_dis}. As seen, we find obvious semantic differences across sub-sets. In the second sub-set (yellow bar), there are more slots related to location (``\textit{train-departure}'' and ``\textit{train-destination}'') while the third sub-set (green bar) mainly involves some slots with numbers, e.g. \textit{restaurant-book people} and \textit{taxi-leave at}.
Most dialogues from the attraction domain are assigned to the second sub-set (blue bar). We conclude that clustering can divide seen data into relatively semantic-independent areas. 
\paragraph{Performs Better Than Using Domain Division}
One may doubt that explicitly dividing data might be better than implicit semantics division by clustering. 
To check this doubt, we construct an explicitly divided baseline according to domains and we train domain-independent experts following its division, where this baseline is named as 
\textbf{DI-Experts}.
For a fair comparison, we average the dialogue vectors in the same domain as the prototype and apply ensemble inference for DI-Expert.
As shown in Figure~\ref{fig:domain_per}, DI-Experts, combining domain-independent experts, shows a significant decrease compared to ours in all domains. 
The reason may be the domain division on seen data focuses on the background of a conversation but ignores the more fine-grained semantics such as user intent, which can be well handled by our cluster method.

\subsection{Analysis on Ensemble Inference}
\label{sec:ensemble}
\begin{figure}[t]
  \centering
  \vspace{-18pt}
  \includegraphics[scale=0.29] {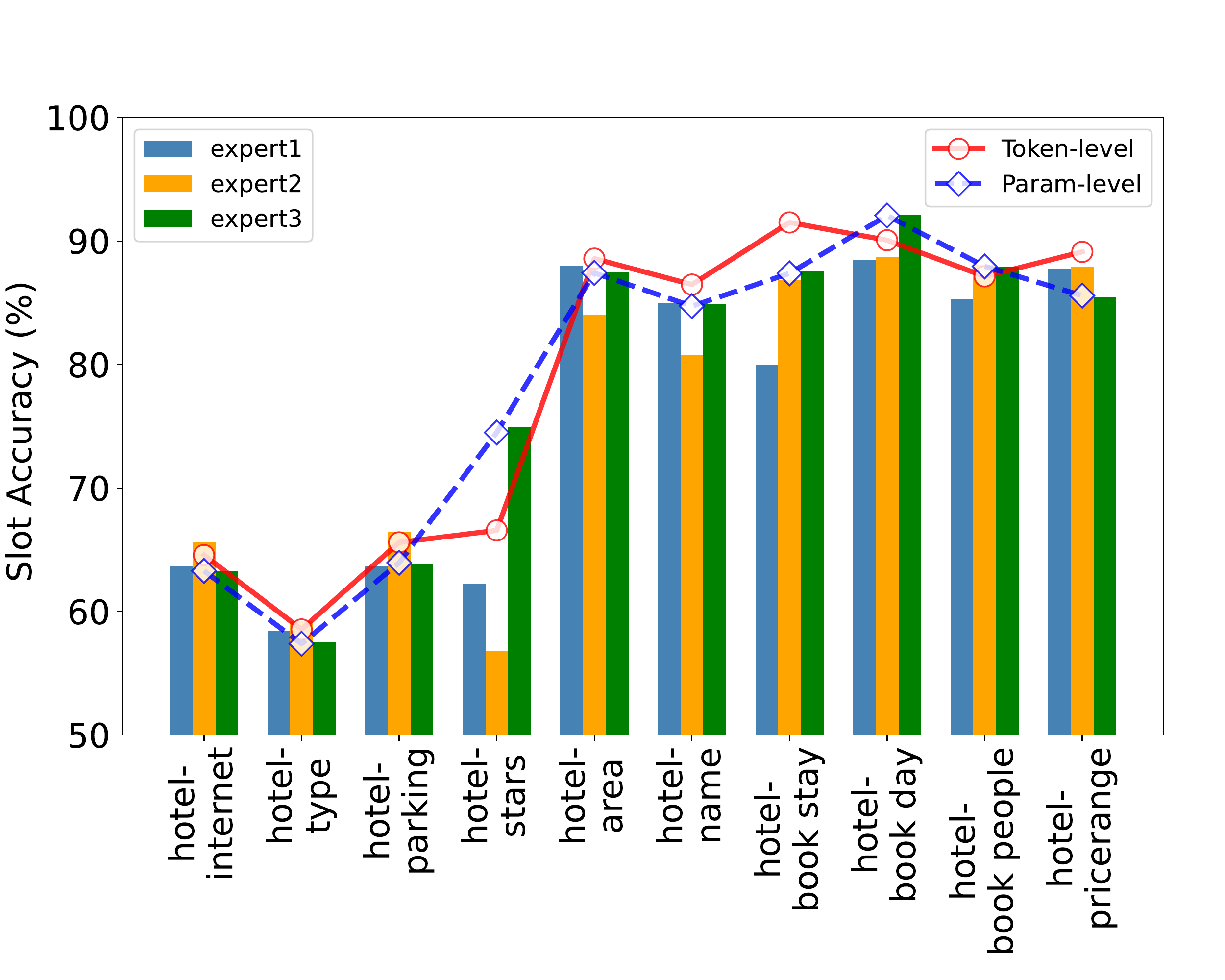}
  \vspace{-10pt}
  \caption{Slot accuracy of different single experts and ensemble models on hotel domain. 
  }
  % \vspace{-10pt}
  \label{fig:hotel_per_slot}
  \end{figure}

\paragraph{Integrates the Advantages of Experts}
Figure \ref{fig:hotel_per_slot} makes a comparison of slot accuracy obtained by ensemble experts and individual experts. 
As shown, 1) the first expert is specialized in ``hotel-area'' and ``hotel-name'' slots, and the third expert performs better on ``hotel-book day'' and ``hotel-book people'', which is consistent with their data-level slot distribution across sub-subsets in Figure~\ref{fig:cluster_dis}, and 2) our ensemble inference methods, especially token-level one, are more accurate, as expected, than the corresponding best expert in most slots, showing the necessity of adopting the ensemble inference.
\paragraph{Requires Lightweight Computational Cost}
Our method requires only tuning and deploying the adapter, which is super lightweight compared to the full pretrained language model training.
Table \ref{tab:resource_com} shows the training and inference overhead in different zero-shot DST models. For a fair comparison, all methods use T5-base as the basic model.
As seen, we only consume 4.9\% parameters compared to the T5-base ``T5DST'' during training, while for inference, our ``Param-level'' and ``Token-level'' only deploy extra +1.6\% and +4.9\% parameters, respectively. The total computing overhead is negligible but we gain significant performance boosts, up to averaging +5.4\% JGA compared to T5-base.
\begin{table}[t]
\centering
\resizebox{\linewidth}{!}{
\begin{tabular}{lrrl}
\toprule
\textbf{Model} & \textbf{Training $\lvert \Theta \rvert$} & \textbf{Inference $\lvert \Theta \rvert$} & \textbf{Average (\%)} \\ \midrule
T5DST                  & 100\%         & 100\%                                       & \underline{37.36}                    \\
T5-Adapter            &  1.6\%            & +1.6\%  & \underline{37.92}                    \\
Ours (Param-level)              & 4.9\%           & +1.6\%                               & \underline{40.76}$^{\Uparrow \textbf{+3.4}}$                    \\
Ours (Token-level)              & 4.9\%           & +4.9\%                                    & \underline{42.71}$^{\Uparrow \textbf{+5.4}}$                    \\ \bottomrule
\end{tabular}}
\caption{Costs for training and inference of methods. $\lvert \Theta \rvert$ denotes the number of trained/ deployed parameters for training and inference, respectively.}
\label{tab:resource_com}
\vspace{-10pt}
\end{table}

\subsection{Complementary to Existing Works}
\label{subsec:complmentary}
Our method for zero-shot DST is a new learning framework, which is expected to complement existing works, e.g. data-level and model-level strategies.
Here we list two representative approaches and show the complementarity.

\paragraph{Data Augmentation Method}
Many methods improve the zero-shot performance and out-of-domain generalization from a data augmentation perspective \citep{Campagna2020ZeroShotTL,Manotumruksa2021ImprovingDS,ding2021rejuvenating,ding2022redistributing}. We train DST using raw data and augmented data from \citet{Campagna2020ZeroShotTL}, respectively, to show further improvement.
As shown in Table \ref{tab:aug_result}, both ``Param-level'' and ``Token-level'' achieve further improvements, i.e. 1.6\% on average, showing the complementarity between ours and the data-level approach.

\begin{table}[t]
\centering
\small
% \resizebox{\linewidth}{!}
{
\begin{tabular}{lll}
\toprule
\textbf{Model}    & \textbf{Raw Data} & \textbf{Augmented Data} \\ \midrule
TRADE            & 19.50                   & 28.30                    \\
Ours (Param-level) & 26.15                  & 27.56$^{\Uparrow \textbf{+1.4}}$                   \\
Ours (Token-level) & 27.71                  & 29.36$^{\Uparrow \textbf{+1.7}}$                   \\ \bottomrule
\end{tabular}}
\caption{Complementarity between ours and data augmentation methods, in terms of zero-shot performance on hotel domain. 
}
% \vspace{-10pt}
\label{tab:aug_result}
\end{table}
\begin{table}[t]
\centering
\small
% \resizebox{\linewidth}{!}
{
\begin{tabular}{llll}
\toprule
\textbf{Model} & \textbf{Attraction} & \textbf{Hotel} & \textbf{Taxi} \\ \midrule
SlotDM     & 36.38               & 25.45          & 67.21               \\
+Our Framework & 37.41$^{\Uparrow \textbf{+1.0}}$                    & 26.58$^{\Uparrow \textbf{+1.1}}$               & 68.02$^{\Uparrow \textbf{+0.8}}$                    \\ \bottomrule
\end{tabular}}
\caption{Complementarity between ours and competitive model-level methods ``SlotDM'', in terms of zero-shot performance on three domains.}
\vspace{-10pt}
\label{tab:slot_dep}
\end{table}

\paragraph{Slot-Slot Dependency Modeling Methods}
Various DST works utilize the correlations among slots and improve the performances on full-shot \citep{Ye2021SlotSD,feng2022dynamic} and zero-shot settings \citep{Wang2022SlotDM}. To benefit from the correlations among slots, we collaborate our framework with ``Slot Prompt Combination'' technique proposed by \citet{Wang2022SlotDM} and observe the zero-shot performance (See Table  \ref{tab:slot_dep}). As shown, our framework could push the SlotDM toward better zero-shot performance by averaging +0.96\% on three domains, demonstrating the complementarity between ours and the model-level approach.

\section{Conclusion}
In this paper, we propose a new learning schema ``divide, conquer, and combine'' to improve the zero-shot generalization in DST. The philosophy behind this is to explicitly divide the seen data into different semantic areas, such disentanglement provides flexibility for mapping the unseen sample to the different experts trained on corresponding semantic areas, and the ensemble results of experts are expected to improve the model generalization.
The experimental results indicate that our model using small trainable parameters reaches state-of-art performances in zero-shot cross-domain DST.

\section*{Limitations}
We conclude the limitations of our schema into two aspects. Firstly, our method benefits from the assumption that there exists similar semantics between the seen data and unseen samples. However, our work might not own obvious advantages in the case where the correlation among domains is weak, such as medical assistant and movie service. But notably, in such cases, most zero-shot learning methods will also fail to show well generalization. 
Secondly, we propose to train semantic-independent DST experts, which is ideal but we believe advanced components could move towards this goal, such as using advanced clustering algorithms and pretrained language models.

\section*{Ethics Statement}
This work does not present any direct ethical issues. We focus on improving the zero-shot cross-domain generalization problem in DST. All experiments are conducted on open datasets and the findings and conclusions of this paper are reported accurately and objectively.

\section*{Acknowledgments}
This work is supported by the National Key Research and Development Program of China (NO.2022YFB3102200) and Strategic Priority Research Program of the Chinese Academy of Sciences with No. XDC02030400. We would like to thank the anonymous reviewers for their valuable comments.

\bibliography{anthology,custom}
\bibliographystyle{acl_natbib}
\newpage
\appendix
\section{Dataset Statistics}
There are 5 domains used in the MultiWOZ dataset in zero-shot settings, which is shown in Table \ref{tab:dataset_mwz}. Additionally, the slot descriptions for all the dialogue state slots are provided in the dataset. The statistics of the SGD dataset are shown in Table \ref{tab:dataset_sgd}
\label{sec:appendix}
\begin{table}[htbp]
 \centering
 \resizebox{\linewidth}{!}{
 \begin{tabular}{|c|l|l|l|l}
 \hline
 
 \multicolumn{1}{l|}{\textbf{Domain}} & \textbf{Slot} & \textit{Train} & \textit{Valid} & \textit{Test} \\ \hline
 
 \multicolumn{1}{c|}{Attraction} & area, name, type & 2717 & 401 & 395 \\ \hline
 
 \multicolumn{1}{c|}{Hotel} & \begin{tabular}[c]{@{}l@{}}area, internet, name, \\ parking, price range, \\ stars, type, book day, \\ book people, book stay\end{tabular} & 3381 & 416 & 394 \\ \hline
 
 \multicolumn{1}{c|}{Restaurant} & \begin{tabular}[c]{@{}l@{}}area, food, name, \\ price range, book day,\\ book people, book time\end{tabular} & 3813 & 438 & 437 \\ \hline
 
 \multicolumn{1}{c|}{Taxi} & \begin{tabular}[c]{@{}l@{}}arriveby, departure, \\ destination, leaveat\end{tabular} & 1654 & 207 & 195 \\ \hline
 
 \multicolumn{1}{c|}{Train} & \begin{tabular}[c]{@{}l@{}}arrive by, day,\\ departure, destination, \\ leaveat, book people\end{tabular} & 3103 & 484 & 494 \\ \hline
 \multicolumn{2}{c|}{Total} & 8438 & 1000 & 1000 \\ \hline
 \end{tabular}}
 \caption{The dataset statistics of MultiWOZ dataset.} 
 \label{tab:dataset_mwz}
 \end{table}

\begin{table}[htbp]
\centering
\resizebox{\linewidth}{!}{
\begin{tabular}{ll|ll}
\hline
\textbf{Domain} & \textbf{\#Dialogs} & \textbf{Domain} & \textbf{\#Dialogs} \\ \hline
Alarm           & 324                & Movies          & 2339               \\
Banks           & 1021               & Music           & 1833               \\
Buses           & 3135               & Payment         & 222                \\
Calendar        & 1602               & RentalCars      & 2510               \\
Events          & 4519               & Restaurants     & 3218               \\
Fights          & 3644               & RideSharing     & 2223               \\
Homes           & 1273               & Services        & 2956               \\
Hotels          & 4992               & Trains          & 350                \\
Media           & 1656               & Travel          & 2808               \\
Messaging       & 298                & Weather         & 1783               \\ \hline
\end{tabular}}
\caption{The dialogues for each domain across train, dev, and test sets in the SGD dataset. The ``Alarm'', ``Messaging'', ``Payment'' and ``Train'' domains are only present in the dev or test sets to test generalization to new domains.}
\label{tab:dataset_sgd}
\end{table}
\iffalse
\section{Additional Experiments}
\subsection{Analysis of Each Expert} 
Figure \ref{fig:per_class} compares the JGA performances obtained by ensemble inference and single experts 
in zero-shot settings. From the results, we observe 
that the token-level ensemble can take good advantage of the performance of each expert, but the parameter-level ensemble tends to be one of the experts.
\begin{figure}[htbp]
  \centering
  \includegraphics[scale=0.23] {hotel_class.pdf}
  \hspace{2mm}
  \includegraphics[scale=0.23] {train_class.pdf}
  \caption{The JGA performances of the ensemble models and single experts on different domains.}
  \label{fig:per_class}
  \end{figure}
\fi
\end{document}